\documentclass[12pt, authoryear]{elsarticle}

\usepackage[edges]{forest}

\usepackage{booktabs}
\usepackage{tabularx}
\usepackage{multirow}
\usepackage{amsmath}

\usepackage{array}
\usepackage{geometry}

\newcommand{\quotes}[1]{``#1''}

\usepackage{tikz}
\usetikzlibrary{shapes.geometric, arrows, positioning}

\usepackage[bookmarks,bookmarksnumbered]{hyperref}
\hypersetup{colorlinks = true,linkcolor = blue,anchorcolor =red,citecolor = blue,filecolor = red,urlcolor = red,
            pdfauthor=author}

\begin{document}
\begin{frontmatter}

    \title{Pivoting Retail Supply Chain with Deep Generative Techniques: Taxonomy, Survey and Insights}
    
    \author[inst1]{Yuan Wang\corref{cor1}}
    \ead{yuan.wang@walmart.com}

    \author[inst2]{Lokesh Kumar Sambasivan}

    \author[inst1]{Mingang Fu}

    \author[inst1]{Prakhar Mehrotra}

    \affiliation[inst1]{organization={Global Technology, Wal-mart},
    addressline={860 W California Ave}, 
    city={Sunnyvale},
    postcode={94086}, 
    state={California},
    country={United States}
    }

    \affiliation[inst2]{organization={Global Technology, Wal-mart},
    addressline={Cessna Business Park, Kadubeesanahalli, ORR}, 
    city={Bengaluru},
    postcode={560087}, 
    state={Karnataka},
    country={India}
    }

    \begin{abstract}
        Generative AI applications, such as ChatGPT or DALL-E, have shown the world their impressive capabilities in generating human-like text or image. Diving deeper, the science stakeholder for those AI applications are Deep Generative Models, a.k.a DGMs, which are designed to learn the underlying distribution of the data and generate new data points that are statistically similar to the original dataset.
        One critical question is raised: how can we leverage DGMs into morden retail supply chain realm? To address this question, this paper expects to provide a comprehensive review of DGMs and discuss their existing and potential usecases in retail supply chain, by (1) providing a taxonomy and overview of state-of-the-art DGMs and their variants, (2) reviewing existing DGM applications in retail supply chain from a end-to-end view of point, and (3) discussing insights and potential directions on how DGMs can be further utilized on solving retail supply chain problems.       
        
    \end{abstract}
\end{frontmatter}

\section{Introduction}

The rapid advancement of artificial intelligence, specifically the blossoming of generative models, has ushered in a new era for diverse fields, fundamentally altering the way we perceive, interact, and innovate with technology. Tools like ChatGPT~\citep{OpenAI2023GPT4TR} and DALL-E~\citep{reddy2021dall} shows the world their incredible power of generating contents such as text or image. 

The impressive capability to generate human-like text, to understand context, and to even create artistic content was nothing short of revolutionary and brought people attention to the science benethes these AI tools --- Deep Generative Models (DGMs). Practitioners are taking actions to attempt leveraging DGMs to pivot their areas as diverse as healthcare~\citep{smith2023evaluating}, finance~\citep{noguer2023generative}, entertainment, and more, in a storming way.   

The retail supply chain is a key component in modern economy, ensuring the smooth flow of goods from origin to end-users. In general, it can be segmented into three distinct phases: purchase, logistics, and sell~\citep{ge2019retail,ayers2017retail}. The purchase phase entails the strategic acquisition of goods from vendors, a process driven by demand forecasting and procurement policies. Logistics, the second phase, involves the transportation of these goods from vendors to distribution centers (DCs), and subsequently from DCs to retail outlets. This step is vital in maintaining the efficiency and effectiveness of the supply chain. The final phase, sell, encompasses the actual sale and subsequent delivery of products to customers, completing the cycle from supplier to consumer.

Due to the complexity of the modern retail supply chain, a host of challenges arise in each phase, motivating researchers and practitioners to seek innovative solutions. 
For example, forecasting demand accurately is a long-existing challenge in the purchase phase, often fraught with the unpredictability of consumer behavior and market conditions~\citep{abolghasemi2020demand}. This challenge is compounded by the necessity to model and simulate the supply chain to anticipate potential disruptions and bottlenecks. The complexity of creating accurate models arises from the numerous variables and stochastic elements inherent in supply chain processes, requiring advanced tools and methodologies~\citep{aamer2020data,babai2022demand}. 
For another instance, the optimality gap —the divergence between the most efficient theoretical solution and the real-world execution—poses a persistent issue. This gap can be attributed to various factors, including but not limited to, logistical stochasticity, supply chain end-to-end visibility, and computational NP-hardness~\citep{chen2017large}. 
These discrepancies often lead to inefficiencies that ripple throughout the supply chain, manifesting in delayed deliveries, increased carrying costs, and diminished service levels~\citep{qu2020optimizing,caro2020future} and therefore it's always impactful and valuable to develop advanced methods to improve the optimality. As a result, the ability to adapt and optimize the end-to-end process is not just a competitive advantage but a survival imperative in the volatile landscape of global commerce. 

With the blossoming of generative AI, some researchers have attempted to provide a survey or review of existing DGMs being used in supply chain problems. Ooi et al provided a comprehensive discussion on potentials of using DGMs accross disciplines including generative AI in retail~\citep{ooi2023potential}. This work is timely and insightful, however, it's more of a high-level discussion of the potentials of DGMs in many areas' supply chain without focusing on retail supply chain. Similar survey can also be found in reference~\citep{sankaran2023generative}.
Article~\citep{wamba2023both} is attempting to address how DGMs applications such as ChatGPT may be a game changer in 21st centenry supply chain excellence. This paper is focusing on specific applications of DGMs, mainly language model, instead of discussing leveraging techniques and methods of DGMs in retail supply chain.
Reference \citep{hosseinnia2023applications} aims to provide a comprehensive vision by reviewing 43 papers on the applications of Deep Learning methods to supply chain management, trends, perspectives, and potential research gaps. Although it contains relevant insights on leveraging deep learning in supply chain, it is not exclusively focused on DGMs or retail supply chain.
Reference \citep{oliveira2023new} is a timely survey which briefly explains recent developments in DGMs, including Energy-based models, Variational Autoencoders, Generative Adversarial Networks, and Autoregressive models. It also reviews recent applications of these techniques to supply chain, although it does not discuss supply chain application in a end-to-end way. This survey is a good starting point for practitioners to understand the landscape of DGMs, however, it's not focused on retail supply chain and lacks insights on how DGMs can be applied to solve specific problems in retail supply chain. 

To our best knowledge, there is no other existing survey or review paper that is focusing on the existing and potential applications of DGMs in retail supply chain. This paper is to fill this gap by providing a comprehensive review of both techniques and business usecases of DGMs in retail supply chain. The contributions of this paper can be summarized as follows:

\begin{itemize}
    \item This paper provides a taxonomy and overview of state-of-the-art DGMs and their variants.
    \item This paper reviewes existing DGM applications in retail supply chain from a end-to-end view of point. 
    \item This paper discusses insights and potential directions on how DGMs can be further utilized on solving retail supply chain problems.
\end{itemize}

The rest of this paper is organized as follows: Section~\ref{sec:DGMs} provides a brief overview of DGMs with taxonomy and introduces serverl popular models and their variants. Section~\ref{sec:app-purchase} discusses the applications of DGMs in the purchase phase. Section~\ref{sec:app-logistics} discusses the applications of DGMs in the logistics phase. Section~\ref{sec:app-sell} discusses the applications of DGMs in the sell phase. And finally, Section~\ref{sec:conclusion} provides a summary of previous three sections and concludes the paper.

\section{Deep Generative Models Overview}\label{sec:DGMs}

Generative modeling palys an increasingly important role in the realm of machine learning, especially when the task revolves around understanding the intricate patterns and structures within data. 
At its core, a generative model seeks to capture the inherent distribution of the data it is trained on, subsequently leveraging this learned distribution to produce or generate novel data samples that are statistically similar to the original dataset~\citep{harshvardhan2020comprehensive}. 
Such models do not just replicate or mimic the input data; instead, they endeavor to unlock the underlying probabilistic mechanisms that might have led to the data to simulate and produce new instances of it.

Contrasting this approach is discriminative modeling. While generative models aim to represent the joint probability distribution of the input and output data, discriminative models focus primarily on the conditional probability of the output given the input~\citep{minka2005discriminative}. Essentially, discriminative models are more concerned with distinguishing between different data classes or categories and making decisions based on these distinctions. For example, in a binary classification task, a discriminative model would focus on determining the boundary that separates the two classes, whereas a generative model would first seek to understand the distribution of each class individually.

Delving deeper into the landscape of generative modeling, the concept of deep generative models (DGMs) is obtaining higher and higher visibilities, with the recent developments of large language models (LLMs). These are a subset of generative models that leverage the deep learning architectures, such as deep neural networks, to capture and reproduce more complex data distributions. The depth in DGMs, afforded by multiple layers of nonlinear processing units, enables them to model intricate patterns and structures in high-dimensional data, often outperforming their shallow counterparts~\citep{xu2015overview}. Their impressive potential and applicability of generativing complex data like images or text, have made DGMs a focal point of contemporary machine learning research and applications.

There are many ways to categorize DGMs according to the their underlying mechanisms, principles of the models and data modality. For example, based on data modality,DGMs can be categorized into image generative models,text generative models, audio generative models, and multi-modal generative models. For another instance, based the network architectures, DGMs can be calssified into forward neural network models, recurrent neural network models, convolutional nerual network, and attention-based nerual networks etc. 

A most common categorization of DGMs is based on learning paradigm, with which DGMs are calssified into two major categories --- explicit density models and implicity density models~\citep{oussidi2018deep,turhan2018recent,regenwetter2022deep}. 

Explicit Density Models (EDMs) specify a functional form for the data distribution (probability density function) to be evaluated explicitly. These models are designed to offers direct insights into the data distribution allowing for exact likelihood calculation, data density estimation, etc. On the other hand, Implicit Density Models (IDMs) do not provide a closed-form expression for the data distribution but can generate samples from it. These are designed to generate new data points that should follow the target distribution without explicitly evaluating the probability of a given data point and to be less restricted by the choice of functional form for the distribution.

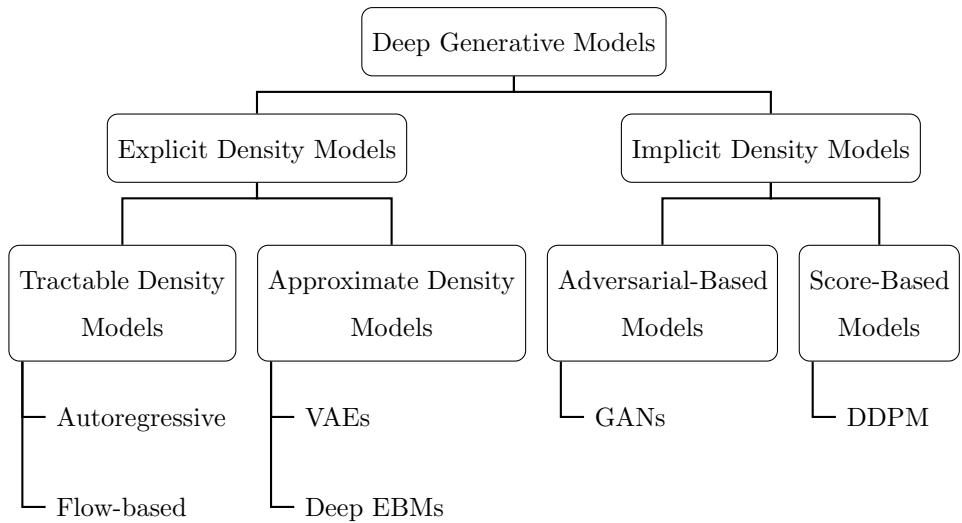
\begin{figure*}
    \centering 
\begin{forest}   
    forked edges,
    for tree={rounded corners, align = center,font=\footnotesize,edge={thick}},
    where={level ()<2}{draw}{for tree={folder,grow'=east}},
    where={level ()==2}{draw}{},
    where={level ()==1}{calign=edge midpoint}{}
    [Deep Generative Models
        [Explicit Density Models
        [Tractable Density \\ Models
            [Autoregressive]
            [Flow-based]
        ]
        [Approximate Density \\ Models
            [VAEs]
            [Deep EBMs]
        ]
        ]
        [Implicit Density Models
        [Adversarial-Based \\ Models
            [GANs]
        ]
        [Score-Based \\ Models
            [DDPM]
        ]
        ]
    ]
\end{forest}
\caption{Taxonomy of Deep Generative Models}\label{taxonomy}
\end{figure*}

According to the different way of estimating the probability likelihood, the EDMs can be further divided into two groups --- tractable density models and approximate density models. IDMs can also be splitted into Adversarial-based and Score-based models based on their different generating processing. Figure~\ref{taxonomy} shows the taxonomy of deep generative models. 

In the following sections, we will briefly introduce serverl popular models (and their variants) in each category to provide readers a high level understanding of their mechanisms.

\subsection{Explicit Density Models}
\subsubsection{Tractable Density Models}
Tractable Density Models (TDMs) are a class of models that can evaluate the probability density function (PDF) of given data points. This is achieved by specifying a functional form $p(x)$ for the data distribution, allowing for exact likelihood calculation, data density estimation, etc. The generation of a new data point is to sample a new data point $x_{new}$ from the distribution $p(x)$, where maximum likelihood estimation is tractable. 

The tractability of these models is a key advantage, enabling them to offer direct insights into the data distribution. However, this tractability comes at the cost of flexibility, as the choice of functional form for the distribution can be restrictive.

There are two major types of TDMs, namely autoregressive models and flow-based models. Autoregressive models use parameterized functions to generate data points sequentially, given all previous ones. The probability of a data point $x$ is given by the product of the conditional probabilities of each element $x_i$ in the data point, conditioned on all previous elements $x_{<i}$, that $p(x)=\prod_{i=1}^{n}p(x_i|x_{<i})$, which is known as the chain rule of probability~\citep{bond2021deep}. 

Flow-based models are designed to learn feature representation (latent variables) with tractable likelihoods. The key idea of flow-based models is to use invertible transformations to map a simple distribution to a complex distribution, allowing for exact likelihood calculation $p_{\theta}(\mathbf{x}) = p(\mathbf{z}) \left| \det\left(J_{f_{\theta}}(\mathbf{x})\right) \right|$, where $\mathbf{z}$ is latent variable. Please refer literatrue~\citep{kobyzev2020normalizing} for more details on flow-based models. The following sections will provide more details on autoregressive models.

There are a few popular autoregressive models including Fully Visible Sigmoid Belief Networks (FVSBNs)~\citep{gan2015learning}, Neural Autoregressive Distribution Estimator (NADE)~\citep{uria2016neural}, and Masked Autoencoder for Distribution Estimation (MADE)~\citep{khajenezhad2020masked}, PixelCNN~\citep{van2016conditional}, PixelRNN~\citep{van2016pixel}, and etc. Each model has its own advantages and disadvantages. For example, FVSBNs are easy to train and sample from, but they are limited to modeling low-dimensional data. NADE is a simple model that can be trained efficiently, but it is limited to modeling low-dimensional data. MADE is a more complex model that can be trained efficiently, but it is limited to modeling low-dimensional data. PixelCNN and PixelRNN are more complex models that can be trained efficiently, but they are limited to modeling low-dimensional data.

An important development in this category is the attention-based  models, which uses self-attention mechanism to capture the long-range dependencies in the data, enabling the method to select the previous steps that are more relevant for the generation processing of the current step. 

When the self-attention mechanism is used with an encoder-decoder nerual network structure, an innovative and impactful architect is developed up with the name of Transformers~\citep{vaswani2017attention}. In Transformers, self-attention mechanism is used to enbale parallelization of the generation process and detect relations between tokens. This is a key efficiency and effectiveness advantage of Transformers over RNNs, which are inherently sequential and cannot be parallelized.

A host of generative variants of Transformers are widely used in the field of natural language processing (NLP) tasks. For example, Bidirectional Encoder Representations from Transformers (BERT)~\citep{devlin2018bert} is an encoder-only Transformers that randomly maskes certain tokens in the input to avoid seeing other tokens and uses bidirectional self-attention to learn contextual representations of words, called embedding.

Another well known variants is called Generative Pre-trained Transformer (GPT)~\citep{radford2018improving}, which is a decoder-only Transformers that uses unidirectional self-attention to autoregressively generate the output sequence by predicting the next token given the previous tokens. ChatGPT, a chat tool being able to generate human-like text, is built on its 3.5 and 4th version of GPT, trained on a massive dataset with as much as 175 billion parameters from 12 transformer layers and 12 attention head.

\subsubsection{Approximate Density Models}
Some other explicit density models often start with an explicit distribution (a prior), but exact inference on this distribution is intractable due to the complexity of the model and data. Thus, they rely on approximate methods to infer the distribution and are called approximate density models (ADMs).

Variational Autoencoders (VAEs)~\citep{kingma2013auto} are a class of generative models that use variational inference to approximate the intractable posterior distribution of the latent variables.

VAEs follows an encoder-decoder structure, where the encoder is used to infer the latent variables from the input data and the decoder is used to generate the output data from the latent variables. During the training process, instead of defining the density function in a tractable way as $p(x) = \prod_{i=1}^{n} p(x_i | x_1, \ldots, x_{i-1})$ in TDMs, VAEs defines the density function in an intractable way $p_\theta (x) = \int p_\theta (z) p_\theta (x | z) dz$, where $z$ is the latent variable. Then VAEs are trained to maximize the evidence lower bound (ELBO) of the data, which is used to approximate the intractable posterior distribution of the latent variables. 

VAEs has its own obvious advantages and disadvantages. On one side, an unique characteristic of VAEs is to encode input data points as a distribution instead of a single point, making the generation process more robust and regularized. However, on the other hand, VAEs tend to be fickle and unstable, as small changes in input data can signficantly change the output~\citep{dosovitskiy2016generating}. 

Overall, VAEs is a popular tool for generating data, especially for images. But recent developments of other DGMs such as adversarial-based or diffusion-based models have made VAEs less popular in the field of image generation.

Another popular ADM is called Deep Energy-Based Models (EBMs) based on the concept of the energy function~\citep{lecun2006tutorial}. In Deep EBMs, the probability density function $p(x)$ can be expressed by an genergy function $E(x)$ that $p(x) = {e^{-E(x)}} / {\int_{\tilde{x} \in X} e^{-E(\tilde{x})}}$. The learned genergy function can associate realistic points with lower values and unrealistic points with higher values. 

The key challenge of training EBMs is to handle the intractable normalization constant in the denominator of the density function. To address this challenge, EBMs use Markov Chain Monte Carlo (MCMC) methods to generate samples from the mdoel distribution, estimate the gradient of the energy function~\citep{song2021train}. For more details and optimize the contrastive divergence~\citep{carreira2005contrastive}. 

There are serveral popular EBMs inlcuding Boltzmann machine~\citep{hinton1983optimal}, Restricted Boltzmann machine (RBM)~\citep{hinton2002training}, and Deep Boltzmann Machines (DBMs)~\citep{salakhutdinov2009deep}. The advantages and disadvantages of EBMs are obvious. EBMs have fewer parameters and more flexible generation process whereas MCMC methods such as random walk and Gibbs sampling makes the training inefficient when applied to high dimensional data.

So far, we introduced four explicit density models, which define the probability density function in a explicit way. The tractability of density function makes them be divided to two categories, namely tractable density models and approximate density models. The summarized density function of these models are as follows:

\begin{itemize}
    \item \textbf{Autoregressive models}
    \begin{align}
        p_{\theta}(x) = \prod_{i=1}^{n}p_{\theta}(x_i|x_{<i})
    \end{align}
    where $x$ is the data point and the density function is tractable. 
    \item \textbf{Flow-based models}
    \begin{align}
        p_{\theta}(\mathbf{x}) = p(\mathbf{z}) \left| \det\left(J_{f_{\theta}}(\mathbf{x})\right) \right|
    \end{align}
    where $\mathbf{z}$ is the latent variable and the density function is tractable.

    \item \textbf{Variational Autoencoders (VAEs)}
    \begin{align}
    p_{\theta}(\mathbf{x}) = \int p(\mathbf{z}) p_{\theta}(\mathbf{x} | \mathbf{z}) d\mathbf{z}
    \end{align}
    where $\mathbf{z}$ is the latent variable and the density function is intractable. To handle the intractability and train the model, ELBO of the data is used to approximate the posterior distribution. 

    \item \textbf{Deep Energy-Based Models (EBMs)}
    \begin{align}
        p_{\theta}(\mathbf{x}) = \frac{e^{-E_{\theta}(\mathbf{x})}}{\int_{\tilde{\mathbf{x}} \in \mathbf{X}} e^{-E_{\theta}(\tilde{\mathbf{x}})}}
    \end{align}
    where $\mathbf{x}$ is the data point and $E_{\theta}(\mathbf{x})$ is the energy function. The density function is intractable and MCMC methods are used to train the model.
\end{itemize}

\subsection{Implicit Density}

Although explicit DGMs shows strong capabilities in many applications, their limitations are also obvious. They either strongly restrict the model architecture to ensure a tractable normalizing constant for likelihood computation or must use surrogate objectives for maximum likelihood approximation. In this section, we will review two ways to circumvent the limitations of explicit DGMs, that are adversarial-based models and score-based models respectively.

\subsubsection{Adversarial-Based Models}
Instead of explicitly defining the probability density function, adversarial-based models (ABMs) use a generator and a discriminator to implicitly discovery the data distribution in an adversarial manner. The generator is to create fake data points as close to the true distribution as possible,  whereas the discriminator is to distinguish between real and fake data points. The training process is to train the generator and discriminator simultaneously, with the generator trying to fool the discriminator and the discriminator trying to distinguish between real and fake data points. A well trained ABM is to have a generator create data points that are indistinguishable between real and fake to the discriminator. A generative adversarial network (GAN) refers to a class of ABMs that uses a neural network as the generator and another neural network as the discriminator.

Literatrues~\citep{gui2021review,creswell2018generative} provide a comprehensive review of GANs and summarized different variants of GANs designed for different purposes. For example, with different type of nerual network, there are Deep Convolutional GANs (DCGAN)~\citep{radford2015unsupervised}, Wasserstein GANs (WGAN)~\citep{arjovsky2017wasserstein}, and etc. With different training tricks, there are Least Squares GANs (LSGAN)~\citep{mao2017least}, Energy-based GANs (EBGAN)~\citep{zhao2016energy}, and etc. With different architectures, there are Conditional GANs (CGAN)~\citep{mirza2014conditional}, Auxiliary Classifier GANs (ACGAN)~\citep{odena2017conditional}, and etc. With different objectives, there are CycleGANs~\citep{zhu2017unpaired}, StarGANs~\citep{choi2020stargan}, and etc. With different applications, there are Text2Image GANs~\citep{reed2016generative}, Image2Image GANs~\citep{isola2017image}, and etc. 

Although all above ABMs showed their impressive performance in different applications, their advantages and disadvantages are also obvious in common. 

On the advantage side, (1) GANs can paralleliz the generation process, which is impossible for autoregressive or VAEs, (2) the generation design is more general with fewer restrictions and is more flexible to various tasks (3) GANs are subjectively thought of be able to produce better quality generations, e.g higher resolution images, than other DGMs~\citep{goodfellow2016nips}. 

On the other side, GANs are often accused of serveral disadvantages: (1) The adversarial nature of GAN makes them difficult and unstable to train. The non-cooperation mechanism cannot guarantee convergence and makes the equilibrium between the generator and discriminator objectives hard to be achieved (2) Another issue is called mode collapse, where the generator produces only a limited set of samples, getting stuck on local optima, and fails to capture the full distribution of the training data (Thanh-Tung and Tran, 2020). (3) GANs are also known to be sensitive to hyperparameters, which can be difficult to tune~\citep{bond2021deep}.

\subsubsection{Score-Based Models}
Another way to circumvent the limitations of explicit DGMs is to use score-based models, which is to model the gradient of the log probability density function, also known as score function, directly with respect to the data and guide the model in generating data that is likely under the learned data distribution~\citep{song2020score}. The score-based models do not have to have a tractable normalizing constant, and can be directly learned by score matching~\citep{vincent2011connection}.

Unlike EBMs which are trained to maximize the log likelihood of the data with intractable denominator $\int_{\tilde{\mathbf{x}} \in \mathbf{X}} e^{-E_{\theta}(\tilde{\mathbf{x}})}$ in their density function, the score-based models use score function instead of density function to sidestep the intractability. The score function of a distribution is defined as $\nabla_{\mathbf{x}} \log p(\mathbf{x})$ and the score-based model is to learn a function such that $s(\mathbf{x}) \approx \nabla_{\mathbf{x}} \log p(\mathbf{x})$. Similar to explicit models, the score-based models can be trained by minimizing the Fisher divergence between the data and model scroes, but without worrying the intractability. A family of training techniques minimizing Fisher divergence are called score matching~\citep{song2020sliced,vincent2011connection}. 
Once the score-based model is trained, an iterative procedure called Langevin dynamics is used to generate sample from it~\citep{parisi1981correlation,grenander1994representations}, which utilize MCMC procedure to generate samples using only the score function. Please refer paper~\citep{song2020score} for more details.

Denoising Diffusion Probabilistic Models (DDPMs) extend this idea by training a sequence of models to reverse a process that gradually adds noise to the data~\citep{ho2020denoising,song2020denoising} to tranform data into a known prior distribution. 
Then, the generation process is the the reverse process called denoising to generate samples from the noise, using stochastic differential equations (SDEs)~\citep{song2020score}. DDPMs have several advantages, including the ability to generate high-quality samples and to perform exact likelihood computation. However, they can be computationally intensive and may require careful tuning of the noise schedule and training process. Please refer paper~\citep{ho2020denoising,song2020denoising} for more details, where they also discussed the detailed connections between DDPMs and score-based models.

Score-based models have achieved impressive performance on many usecases, especially in image generation~\citep{song2019generative,song2020improved,ho2022cascaded} (outperform GANs~\citep{dhariwal2021diffusion}), audio synthesis~\citep{chen2020wavegrad,kong2020diffwave,popov2021grad} and audio generation~\citep{cai2020learning}. 

\section{Applications in Purchase Phase}\label{sec:app-purchase}
In this section, we will review existing applications of DGMs in the purchase phase, such as deman forecasting, procurement policy optimization, and etc. Some potential applications will also be discussed.
\subsection{Demand Forecasting}
A comprehensive survey provided a comprehensive review for using transformer in time series forecasting~\citep{wen2022transformers}. This paper proposed a new taxonomy consisting of network design and application, summarized serverl representative methods in each category and discussed their strengths and limitations by experimental evaluation, showing performance improvements of transformer over traditional methods. 
However, literatrue~\citep{zeng2205transformers} questions the validity of this line of research and presents experimental results on the effectiveness of transformer models for time-series forecasting on nine real-life datasets, claiming that transformer models are not as effective as people expect. 

Specifically in the context of retail supply chain, literatrue~\citep{valles2022approaching} discusses leveraging transformer-based models to forecast demand. It developes three alternatives to tackle the problem of forecast sales at day/store/item level. The empirical results show that the proposed models outperform the baseline models in terms of accuracy and efficiency, by using sipme sequence to sequence architecture with minimal data processing effort. In addition, it describes a training trick to make the model time independent and general.
Reference~\citep{joshaghani2023retail} proposed an meta-learning framework based on deep encoder-decoder nerual networks to forecast demand for retail supply chain. The method is designed to automatically extracts metafeatures from raw time series and exogenous variables, where encoder-decoder temporal convolutional networks is to capture time series features and encoder-decoder attention LSTM is adopted to capture higher representation features of exogenous retail variables. 
Paper \citep{zhang2023intermittent} introduces transformer-based models to forecast intermittent demand and tested empirically on a dataset with 925 intermittent demand items, compared with two traditional methods such as Croston’s and the Syntetos–Boylan approximation as well as several popular neural network architectures including feedforward neural networks, recurrent neural networks, and long short-term memory. The results show that Transformer performs very well against other methods in a varity of settings. Similar usage of transformer in other demand forecasting areas can also be found in finance~\citep{gokhale2023transfer}, energy~\citep{koohfar2023prediction}, aviation~\citep{wang2022flight}, tourism~\citep{bi2023tourism}.

\subsection{Merchandising Planning}
In retail, merchandising planning is to determine the product assortment, pricing, and inventory levels for each product category to ensure that the right products are available at the right time and right place to meet the demand of customers. In general, its tasks consist of market analysis, product selection, supplier selection, inventory planning and etc. 

A key step of merchandising planning is the market research to understand the market trends and customer needs: what they buy, when, where, why and how often. The overall market analysis is huge project but DGMs can help on serveral specific subtasks such as market research, trend analysis, and etc that LLMs~(please refer~\citep{naveed2023comprehensive} for a comprehensive review of LLMs) can be used to help retailers to understand customer's perception of products and brands by analyzing the reviews and comments from customers. For example, paper~\citep{li2023language} explored and accessed if and when appropriate adaptations of these models can substitute for market research using human surveys, focusing on perception analysis, demonstrating LLMs is a reliable augmentation or even substitute for human brand perception surveys. A similar marketing analysis usecases are also seen from paper~\citep{thida2023automated} for job market and paper~\citep{deng2023llms} for financial market.

Another key task of merchandising planning is to select suppliers, a process of finding, evaluating, and engaging suppliers for acquiring goods and services. DGMs can also help on this process. For example, supplier selection in genearl is an optimization problem aiming to achieve the best trade-off between cost, quality, and delivery stability etc. One typical challenge is the high numbers of decision attributes and low numbers of data samples for decision analysis. Paper~\citep{lin2022innovative} proposed a dynamic supply chain member selection algorithm based on conditional generative adversarial networks (CGANs), which analyzes the index data of the suppliers' relevant factors. 

\subsection{Invertory Allocation}
Inventory allocation is to determine the optimal assortment of products to be stocked at each location to maximize the customer experience and subsequently the profit. 

To our best knowledge, we haven't found any existing applications of DGMs in inventory allocation and replenishment. However, this paper believes some potential problems can be solved by DGMs. For example, in the retail context, items are often sold in an associated way and the demand of one item is often correlated with the demand of other items. One famous example is the demand of beer is correlated with the demand of diapers. This correlation is called cross-selling. Another key factor needs to be considered in inventory allocation is called substitution effect, which is to consider the demand of one item can be substituted by the demand of other items. Both cross-selling and substitution effect requires to understand the correlations between items, namely the joint distribution of the demand of all items.

Understanding the joint distributions for milliions of SKUs is not the trivial task with traditional machine learning methods. This paper believes DGMs can hopefully used in this task, as DMGs is well known for its capabilities of modeling the joint distribution of multiple variables. 

\subsection{Replenishment Optimization}
Inventory replenishment is to determine the optimal quantity of each product to be ordered at each location to meet the demand, aimining to find the right balance between the cost of holding inventory and the cost of not having enough inventory to meet the demand.

The key challenge is to find an optimal expection trade-off between inventory holding cost and sales opportunity cost, given the variabilities of demand, replenishment lead time, fulfullment rate etc. To address the stochasticity, replenishment probelms are usually solved with two steps: (1) prediction of demand and other stochastic variables; (2) optimization of replenishment policy.
Unlike this prediction-then-optimization structure, paper~\citep{Qi2023} proposed a one-step end-to-end (E2E) framework that uses deep learning models to output the suggested replenishment amount directly from input features without any intermediate step. In its nerual network architecture, an autoregressive multi-quantile RNN (MQRNN) is used to receives multiple time series (demand or promotion series) and other fully connected layers are used to generate optimal results. The experimental results show that the proposed E2E framework outperforms the traditional methods in terms of inventory cost and service level. 

DGMs are also widely utilized together with deep reinforcement learning techniques for inventory management problems, where DGMs naturally take the role to learn the stochasticity of RL rewards. 
Paper~\citep{sekar2022reinforcement} provides an overview of reinforcement learning algorithms used in inventory control, highlighting that value-based methods often utilize neural networks (including DGMs) to approximate the optimal action-value functions and are most fruitful when data is scarce or gathering new data is difficult.
Study~\citep{andaz2023learning} discussed  incorporating a deep generative model for the arrivals process as part of the history replay, with which reinforcement learning is applied to optimize the replenishment policy. By formulating the problem as an exogenous decision process, this paper can obtain a reduction to supervised learning. 
Literatrue~\citep{dehaybe2023deep} discusses the growing interest in deep reinforcement learning approaches to inventory control problems, particularly those characterized by features and dynamics too complex for classical dynamic or mathematical programming, also calling the capabilities of nerual networks (including DGMs) to approximate the optimal action-value functions. 
Paper~\citep{Park2023} focuses on adaptive inventory replenishment using a structured reinforcement learning algorithm, particularly effective when demand distributions are non-stationary and change over time. 
To address the stochasticity, DGMs are widely utilized together with deep reinforcement learning techniques. Please also refer papers~\citep{deng2021deep,madeka2022deep} for readers interested in this topic.

\section{Applications in Logistics Phase}\label{sec:app-logistics}

In this section, we will review existing applications of DGMs in the logistics phase, such as network planning, warehouse operations optimization, transportation optimization, and transportation vehicle routing. Some potential applications will also be discussed.

\subsection{Discrete Event Simulation}
Discrete Event Simulation (DES) is a powerful tool for modeling and analyzing complex systems. It is a popular tool in supply chain management, especially in the logistics phase. For example, DES is used to model and analyze the operations of the logistics entities to find insightful solutions to improve the efficiency and effectiveness of the supply chain. 
The key idea of DES is to model the system operations as a set of discrete events and each event is excuted at a specific point in time as th-occurance. The brief proces of DES is as follows. Given a sequence of events ordered chronologically, the system state is updated (proceeding the simulation clock) by executing the earliest event. Then, new events are generated based on all previous executed events and put into the event queue. The simulation process is repeated until the system reaches the end condition. Finally, the output of DES is a sequence of excuted events with their time to be excuted.

Intuitively, the event generating process is similar to the autoregressive generation process of text, where the next token is generated based on all previous tokens. This similarity inspires researchers to leverage DGMs to model the event generating process in DES, given the fact that large retailers always own a tons of event data that can be used to the training of DGMs.

Although, we haven't found exact applications of using DGMs in DES, there are some related work are very similar to this idea. For examples, a close research track is called temporal point processes (TPPs). TPPs are a class of stochastic models that describe the occurrence of events over time, given the probability distribution of time intervals and types of future events occurrence conditioned on the impacts of historical observations.  

Following this topic, literatrue~\citep{shchur2021neural} provided extensive review on using deep learning techniques for TPPs. Futher,paper~\citep{lin2022exploring} designed a framework of generative neural temporal point process (GNTPP) by employing the deep generative models as the probabilistic decoder to approximate the target distribution of occurrence events. The paper also conducted experiments to study the effectiveness of DGMs in TPPs, showing that GNTPP can improve the predictive performance of TPPs, and demonstrates its good fitting ability. 
GNTPP established two tasks: (1) historical encoder to learn how sequential historical evnets (point time and type) will impact next evnet, (2) probabilistic decoder is set up conditional probabily of the next event with parameters obtained by historical encoder. Several DGMs can be found in each part. For example, reviesed attentive model~\citep{zuo2020transformer,zhang2021learning} in historical encoder, VAEs~\citep{pan2020variational} and GAN~\citep{xiao2017wasserstein} in probabilistic decoder.

Although DES is not exactly the same as TPPs and has its unique challenges, this paper holds an optimistic view on the potential of using DGMs in discrete event simulation.

\subsection{Vehicle Routing Optimization}
Veihecle routing is a classic problem in logistics, which is to find the optimal route and tour for a fleet of vehicles to serve a set of customers. Many DGMs researchers treat the routing process as the generation of next location given the current location, which is similar to the autoregressive generation process of text. This similarity inspires researchers to leverage DGMs to model the routing process.

There are tons of studies exploring the possibility of utilizing Transformer to solve VRPs. 
The survey~\citep{bogyrbayeva2022learning} provides an overview of machine learning methods applied to solve vehicle routing problems, including discussions on transformer models and attention mechanisms given their prevalence in recent machine learning research. 
Specifically for DGMs, the work~\citep{kool2018attention} is an early study using a standard transformer to encode the cities and the decoding is sequential with a query composed of the first city, the last city in the partial tour and a global representation of all cities. Training is carried out with reinforce and a deterministic baseline. Later, The work~\citep{Bresson2021} focused on the architecture of leveraging Transformer architecture for Traveler Salesman Problem (TSP) and observed that the Transformer architecture can be successful to solve the TSP expanding the success of Transformer for natural language process and computer vision. 
In addtion, reference~\citep{peng2020deep} introduces a dynamic attention model with a dynamic encoder-decoder architecture, which could dynamically explore node features and exploit hidden structural information effectively, focusing on vehicle routing problems.
In paper~\citep{Rabecq2022}, a general deep reinforcement learning framework is introduced to solve routing problems. An encoder based on an improved Graph Attention Network (GAT) is used, which forms a graph-attention model with a Transformer decoder. This paper employs two deep reinforcement learning algorithms to train the model.
An improved transformer model with both multi-head attention mechanism and attention to attention mechanism (AOA) is proposed in paper~\citep{Lei2022}, where the model is applied to solve the low-carbon multi-depot vehicle routing problem (MDVRP), demonstrating the versatility of transformer models in addressing various specifications of the routing problems. 
Paper~\citep{Tang2023} discusses the use of a Graph Transformer combined with reinforcement learning to address the classic combinatorial optimization problem. 
Article~\citep{Fellek2023} tackles the Capacitated Vehicle Routing Problem with Time Windows (CVRPTW) and the Capacitated Pickup and Delivery Problem with Time Windows (CPDPTW) by developing a deep learning algorithm with an Attention Encoder-Decoder structure and a novel insertion heuristic for the feasibility check of CPDPTW problem. 

The potential capabilities of DGMs in VRPs is still in a strong debt. Paper~\citep{wu2022learning} pointed out that tranditional methods outperformed transformer-based models for VRPs in 2021 Amazon Last Mile Routing Challenge dataset~\citep{merchan20222021}. The paper explained as the transformer-based models require a large amount of data to train, which is not available in the challenge. 

\subsection{Logistics Network optimization}
In the retail supply chian, network optimization consists of multiple key tasks, such as (1) facility location to determine optimal number, location, and DC-store assignment of DCs to cover the demand of customers; (2) transportation network design to determine the optimal transportation routes and modes for moving loads from DCs to stores; (3) service area optimization to determine the optimal catchment area of each store aiming to find a proper balance between service level agreement and transportation cost.

Those problems are long existing and well studied in the literature that most of the existing methods are based on mathematical programming. Neural networks are also widely used for solving network planning problems, although they are not exact DGMs. 

For example, paper~\citep{liu2023end} approximated the matching-based multi-objective facility location problem as node and edge classfication or prediction tasks on graphs. Then it used graph convolutional networks (GCNs) to extract high-dimensional characteristic of grahs and learn optimal policies to solve the problem instead of relying on heuristics methods. In fact, this research track is within the scope of leveraging graph neural network (GNN) to solve combinatorial optimization problems. Please refer literatrue~\citep{joshi2019efficient,joshi2019learning,khalil2017learning} for readers interested in this topic. 

Another widely studied problem in network planning is the transportation routing planning. Its task is to plan and design the optimal routes and modes between nodes in a transportation network, aiming at minimizing the transportation cost while satisfying the Service Level Agreement (SLA). The common methods are based on mixed integer programming, consisting of two steps: (1) generating valid candidate routes and modes; (2) selecting the optimal routes and modes to cover origantion-destination demands, a.k.a the set covering problem (SCP). When the problem scale is growing, the problem can be expolentially difficult to solve. To reduce the amount of candidate routes and modes, column generation technique is often used. There are a set of studies considering using neural network to approximate SCP and reduce the computational complexity as well. 
For example, paper~\citep{shafi2023graph} proposed the Graph-SCP to accelerate solving SCP with GNNs. This paper cast  an instance of SCP as a graph and learn a graph neural network (GNN) to predict a subgraph that contains the solution with offline training. The results showed that Graph-SCP achieves between 30--70 percent reduction of input problem size which leads to run times improvements of up to 25x, compared to the commercial solvers. This research track also belongs to the scope of leveraging GNNs to solve combinatorial optimization problems. 

To our best knowledge, there is no exact studies directly using DGMs to solve either facility location problem or routing planning problem. A comprehensive reivew can be seen in~\citep{Wang2021}, which discussd deep reinforcement learning in network optimzation problems. However, this paper still holds an optimistic view on the potential of using DGMs in network planning, such as incooperating DGMs with reinforcement learning techniques for problems. 



\subsection{Estimated Time of Arrival}
Providing customers or stores accurate time of delivery is of great importance to their purchasing decision and post-purchase experiences. Estimated Time of Arrival (ETA) is a well defined and widely studied problem and a variety of prediction algorithms are utilized for vairous operations scenarios. For example, literatrue~\citep{AlNaim2021} reviewed and compared different algorithms for ETA using geospatial transportation data. Study~\citep{reich2019survey} prodived the survey of ETA prediction methods in public transport networks. In the context of retail supply chain, particularly in last mile delivery page, there are some unique challenges.

First, the overall ETA is usally the sum of a sequence of correlated sub-ETAs. For example, to fulfill the last mile delivery in the omnichannel retail, it consists of multiple steps: picking, sorting, requesting drivers, loading, transporting and delivering. These steps are often sequentially dependent that variation in previous steps will affect distribution of later steps. For instance, transportation time tends to be shorter if delaying in loading has happened. This characteristic of sequentially dependency makes the ETA more difficult to be accurate, as we are not only estimating the distribution in each single step but also approximating their joint distributions. 
In this case, autoregressive DGMs are straightforward to be applied to model the joint distribution of multiple correlated sub-ETAs. Paper~\citep{Zhou2023} proposed inductive graph Transformer algorithm for delivery time prediction, which uses Temporal and Heterogeneous GCN to encode high-order semantic information of orders into latent element embeddings and then applied a customized transformer for the ETA predcition. 

Other than the above studies, Sun et proposed a novel, brief and effective framework mainly based on feed-forward network (FFN) for ETA, FFN with Multifactor Attention (FMA-ETA), where a novel Multi-factor Attention mechanism is utilized to deal with different category features and aggregate the information purposefully~\citep{Sun2021}. Similar applications of attion mechanisms can be seen from paper~\citep{Reich2022} where a combination of an attention mechanism with a dynamically changing recurrent neural network (RNN)-based encoder library is used for public transport ETA prediction.

\section{Applications in Sell Phase}\label{sec:app-sell}

\subsection{Customer Service and Engagement}
Customer service and engagement is a key operation in sell phase to provide customers with the best service and experience to increase the conversion rate. With an intuitive understanding, customer service and engagement is a process of communication between customers and retailers. The communication can be in various forms, such as text, voice, image, video, and etc.
Tranditional customer service and engagement is often done by human in the call center, which is time consuming and costly. With the development of DGMs, especially LLMs, the customer service and engagement can be automated and improved. 

Utilizing LLMs for customer service is straightforward and natural, as LLMs are well known for their capabilities of natural language generation. From the engineering practice point of view, there are four major components for a LLM-based automated customer service system: data collection and preprocessing; knowledge embedding; model selection;system integration.

(1) Data collection and preprocessing is to collect and preprocess the data for training or fine tuning the LLMs, including text, voice, image, video, and etc. The data can be preprocessed by removing the sensitive information, such as customer name, address, phone number, and etc. The data can also be augmented by adding noise, such as typos, grammatical errors, and etc.
(2) Knowledge embedding is to convert the collected data or knowledge into vector representations in the mathematical space. The embedding is to provide the LLMs contextual information for the converstaion so that models can provide meaningful response to user queries. Paper~\citep{su2022one} introduced a instruction-finetuned text embedding method aiming at any task. 
(3) Model selection is to select the appropriate LLM for the specific automation task. This key idea is to find a proper balance between model performance and inference cost~\citep{chung2022scaling}. (4) And last step is to integrate the selected model into the system and deploy it to the production environment. Thanks to the rapid development of LLMs application, there are a lot of open source projects and tools available for the integration and deployment, such as langchain~\citep{topsakal2023creating} and gradio~\citep{abid2019gradio}. 

\subsection{Search and Recommendation}

Search and recommendation is to provide customers with the most relevant products based on their search queries or browsing history. It is a one of most critical tasks in the sell phase for helping customers find the products they want and increase the conversion rate. Recent development of DGMs, especially LLMs, will certainly have a huge impact and pivots the research direction of search and recommendation. For example, the tranditional interface is by key words, where customer input key words such as \quotes{XX phone} or \quotes{YY laptop} and returns the most relevant products. In the era of LLMs, the input can be comprehensive and conversational such as \quotes{I need a list of items for my upcoming road trip}. This is a huge improvement and will certainly improve the search accuracy and customer experience.

Paper~\citep{li2023large} provides a timely survey and visionary discussion for generative recommendation with LLMs. In this paper, it divides the generaive recommendation into two major steps (1) ID creation (2) Generative recommendation. 

ID creation is to create an ID for each item such as a sequence of tokens that can uniquely identify an entity. To implenment generative recommendation with LLMs,the input (particularly user and item IDs) should be made into the right format that is compatible with LLMs. The key idea is to use a small amount of tokens to distinguish as well as effectivelya and accurately distinguish an astronomical number of items. This process is similary to embedding the items into a lower representation space. 
There are a few existing ways to create IDs. For instance, 
Singular Value Decomposition acquires an items ID tokens from its latent factors~\citep{Petrov2023}. A noise-adding operation can ensure that there are no identical item embeddings, and thus make each item ID unique. 
Another way is called product quantization (PQ). Paper~\citep{hou2023learning} proposes VQ-Rec, a novel approach to learning Vector-Quantized item representations for transferable sequential Recommenders based on the idea of product quantization~\citep{jegou2010product}.
Other ID creation methods can be seen from colloerative indexing~\citep{hua2023index}.

After IDs are created, the next step is to generate recommendations with LLMs. Given the prompt, the recommendation process is basiccally to generate outputs (predicted item IDs) in an autoregressive way like natural language generation. The recommendation process is often involved with leveraging LLMs for following tasks such as rating predcition~\citep{Liu2023}, top-N recommendation~\citep{Wang2022,zhang2023recommendation}, sequential recommendation~\citep{Xu2023,Petrov2023}, explainable recommendation~\citep{Li2023,Cui2022}, review summarization~\citep{Liu2023a,Wang2022a} and converstaional recommendation~\citep{Friedman2023,Gao2023}. For more information, please refer to paper~\citep{li2023large}.

Overall, generative search and recommendation is a hot and promising research direction and we will see more and more theoretical advancements as well as practical applications in this area in a short future.





\section{Discussion}\label{sec:conclusion}

This paper provides a comprehensive review of DGMs, and its existing and potential applications in retail supply chain. We first introduced the definition of DGMs, its taxonomy, and some popular models. Then, we reviewed the existing applications of DGMs in the purchase phase, logistics phase, and sell phase, while discussed our insights for their the potentials.

Overall, from the model capability perspective, all DGMs showed their impressive potentials in the area of retail supply chain. We belive DGMs are powerful tools for the following two scenarios: 
\begin{itemize}
    \item For \textbf{predictive} tasks. DGMs showed superier performance in modeling the joint distribution or sequential dependency of multiple variables, such as demand forecast, ETA and so on.
    \item For \textbf{prescriptive} tasks. DGMs showed greate potentials for stochastic optimization, by incooperating with reinforcement learning techniques for problems with sequential decisions, such as inventory replenishment, routing planning and so on.
\end{itemize}
However, other than that, we haven't seen much obvious advantages of DGMs over traditional methods for the deterministic optimization problems, such as staff scheduling optimization or network planning. 

In addtion, some relative mature applications such LLMs have been showing its pivotal role in the retail supply chain. For example, it's increasingly promising to utilize LLMs for customer service and engagement, search and recommendation, and etc. We believe LLMs will continue to play a key role in the retail supply chain in the near future.

DGMs are still in its early stage and there are a lot of challenges and opportunities. We believe DGMs will have a huge impact on the retail supply chain in the near future.

\bibliographystyle{elsarticle-harv} 
\bibliography{ref}
\end{document}